%% file: main.tex
\documentclass[runningheads]{llncs}

\usepackage[mobile]{eccv}

\usepackage{eccvabbrv}

\usepackage{graphicx}
\usepackage{booktabs}

\usepackage[accsupp]{axessibility}  

\usepackage{hyperref}

\usepackage{orcidlink}

\usepackage{bm}
\usepackage{amssymb}
\usepackage{mathtools}
\usepackage{indentfirst}

\usepackage{colortbl}  

\begin{document}

\title{Occam's Razor: Internalizing Temporal Consistency in Video Object-Centric Learning without Explicit Regularization}

\titlerunning{xSSC}

\author{
Rongzhen Zhao\inst{1}\orcidlink{0009-0000-3964-7336} \and
Zhiyuan Li\inst{1}\orcidlink{0000-0002-1804-3485} \and
Juho Kannala\inst{2,3}\orcidlink{0000-0001-5088-4041} \and
Joni Pajarinen\inst{1}\orcidlink{0000-0003-4469-8191}
}

\authorrunning{R.~Zhao et al.}

\institute{
Department of Electrical Engineering and Automation, Aalto University, Finland \email{rongzhen.zhao@aalto.fi} \and
Department of Computer Science, Aalto University, Finland \and
Center for Machine Vision and Signal Analysis, University of Oulu, Finland
}

\maketitle

\begin{abstract}
Video Object-Centric Learning (OCL) aims to represent objects as \textit{slot} vectors and maintain their consistency across frames. Slot-Slot Contrastive (SSC) loss has become the cornerstone for state-of-the-art (SOTA) video OCL methods. While highly effective, SSC relies on one-to-one object correspondence across frames and introduces an extra loss. Following Occam's Razor, we propose a paradigm shift: temporal consistency is better enforced as an implicit model design rather than an explicit loss.
To elegantly exclude SSC (\textbf{xSSC}), we introduce two quasi-zero-overhead synergistic mechanisms:
(\textit{i}) Chrono-Channel Decomposition (CCD) structurally disentangles slot representations along the channel dimension into \textit{static} and \textit{dynamic} sub-spaces, serving as an empirically unified information bottleneck;
(\textit{ii}) Cross-Temporal Reconstruction (CTR) stochastically reconstructs target features of either the current or previous time step by fusing current slots' static channels and target slots' dynamic channels, using a single standard OCL decoder with minor training adaptation.
Thereby, the slot sets inherently learn temporal consistency by minimizing the standard reconstruction error alone.
Extensive experiments show that integrating xSSC into leading baselines not only improves training efficiency but also establishes new SOTAs on video object discovery and recognition tasks. Furthermore, our PCA and gradient analyses confirm that objects' time-invariant semantics and time-variant kinematics are encoded into the proposed sub-spaces.
Our source code, model checkpoints and training logs are provided on https://github.com/Genera1Z/xSSC.

\end{abstract}

\section{Introduction}
\label{sect:introduction}

With unsupervised video Object-Centric Learning (OCL) \cite{locatello2020slotatt, kipf2022savi}, objects and the background in a video can be represented as \textit{slot}s, where each vector corresponds to an object, and maintained consistent across continuous frames.
Such object-level representations provide a structured and interpretable abstraction of dynamic scenes, which has been shown to benefit downstream tasks including prediction, reasoning, dynamics modeling, planning, and decision-making \cite{wu2022slotformer}. Moreover, such representation learning paradigm is also cognitively plausible, as it aligns with how humans perceive and track entities over time \cite{cavanagh2011visual}.

In recent years, the Slot-Slot Contrastive (SSC) loss \cite{manasyan2025slotcontrast} has significantly advanced state-of-the-art (SOTA) video OCL architectures.
By elegantly utilizing objective InfoNCE \cite{oord2018infonce} to pull together slots representing the same object across frames while pushing away different objects, SSC has established a high standard for temporal consistency in video OCL. Its robust explicit regularization has deeply influenced a lineage of strong SOTAs \cite{manasyan2025slotcontrast, zhao2025randsfq, zhao2025smoothsa}.

Despite its undeniable empirical success, relying on explicit contrastive matching inherently assumes a relatively stable one-to-one object correspondence across adjacent frames. In highly dynamic visual scenes where objects frequently occlude each other, disappear and reappear, such as in the MOVi-series datasets \cite{greff2022kubric}, these explicit constraints can become brittle bottlenecks.
Previous non-contrastive attempts, such as VideoSAUR's temporal similarity loss \cite{zadaianchuk2024videosaur}, suffer from similar narrow assumptions that object patches remain constantly visible in the subsequent frame.
Furthermore, appending auxiliary explicit loss terms inevitably introduces hyperparameters and overhead \cite{zadaianchuk2024videosaur, manasyan2025slotcontrast}.

\begin{figure}[t]
\centering
\includegraphics[width=\linewidth]{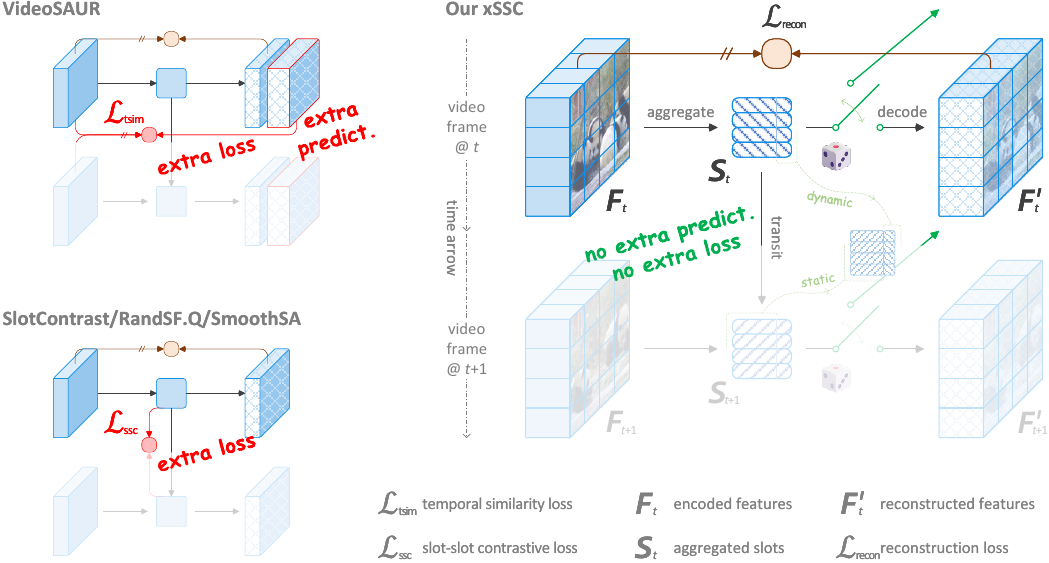}
\caption{
Learning temporal consistency for video Object-Centric Learning (OCL) via \textcolor[RGB]{0,224,80}{implicit model design} instead of \textcolor{red}{explicit loss term}.
Mainstream video OCL methods adopt the encode-aggregate/transit-decode architecture \cite{zhao2025vvo}, with minimizing the difference $\mathcal{L}_\mathrm{recon}$ between encoded features $\bm{F}_t$ and decoded reconstruction $\bm{F}'_t$ as the primary supervision.
(\textit{left}) The temporal consistency of slots $\{ \bm{S}_t \}_t$ is explicitly enforced (\Cref{sect:preliminary}) by temporal similarity loss $\mathcal{L}_\mathrm{tsim}$ in VideoSAUR \cite{zadaianchuk2024videosaur}, or by slot-slot contrastive loss $\mathcal{L}_\mathrm{ssc}$ in SlotContrast \cite{manasyan2025slotcontrast} and its followers \cite{zhao2025randsfq, zhao2025smoothsa}, causing extra overhead and hyper-parameters though.
(\textit{right}) We implicitly learn temporal consistency in our xSSC by randomly forcing either the \textcolor[RGB]{0,224,80}{static} part of future slots $\bm{S}_{t+1}$ or current slots $\bm{S}_t$, concatenated with the \textcolor[RGB]{0,224,80}{dynamic} part of the current slots, to reconstruction current features $\bm{F}_t$, avoiding any extra prediction or loss (\Cref{sect:ccd,sect:ctr,sect:objective}).
}
\label{fig:teaser}
\end{figure}

As shown in \Cref{fig:teaser}, \textbf{we propose a paradigm shift}: temporal consistency is better enforced as an implicit model design rather than an explicit loss. We introduce xSSC (Excluding SSC) , demonstrating that the powerful temporal consistency pioneered by explicit InfoNCE regularization can be elegantly and completely internalized through two synergistic mechanisms:
\begin{itemize}
\item Chrono-Channel Decomposition (CCD) physically disentangles the slot representations along the channel dimension into a static part (encoding time-invariant object identity and appearance) and a dynamic part (encoding time-variant kinematic state and motion). Through extensive empirical investigation, we discover a unified optimal information bottleneck: strictly allocating a compact subspace (one-quarter of the total channels) to the dynamic features acts as a powerful regularizer across diverse datasets. This ratio aligns perfectly with the physical intuition that an object's changing kinematic state operates on a much lower-dimensional manifold compared to its rich, persistent appearance.
\item Cross-Temporal Reconstruction (CTR) introduces a parameter-free temporal reconstruction task, to enforce this decomposition without explicit contrastive penalties. During training, the model must stochastically reconstruct features of either the current or the previous time step. Crucially, this is achieved by fusing the static channels of the current frame with the dynamic channels of the chosen target frame. Because the network is forced to use the current static channels to anchor identity across different time steps, it mathematically prevents the leakage of static features into the dynamic subspace. Furthermore, CTFB operates through a single, standard OCL decoding head. Only minor yet quasi-zero-overhead adaptations are needed for training, while the inference is nothing more than a standard decoding.
\end{itemize}

Extensive experiments demonstrate that xSSC not only achieves higher training efficiency by dropping the SSC loss but also sets SOTA on highly dynamic video object discovery and recognition tasks. To rigorously validate our structural claims, we provide comprehensive explainability analyses. Temporal PCA trajectories and downstream MLP gradient attributions confirm that xSSC successfully disentangles objects' time-invariant semantics and time-variant kinematics into our proposed sub-spaces.

In summary, our key contributions are:
\begin{itemize}
\item \textbf{A Paradigm Shift from Explicit Loss to Implicit Design}: We propose xSSC, demonstrating that explicit temporal penalties like SSC can be elegantly and completely excluded from SOTA video OCL methods, adhering to Occam's Razor.
\item \textbf{Novel Synergistic Mechanisms}: We introduce Chrono-Channel Decomposition (CCD) as an empirically unified information bottleneck, and Cross-Temporal Reconstruction (CTR). Together, they inherently enforce temporal consistency using typical OCL decoding with zero overhead.
\item \textbf{SOTA Performance and Interpretability}: Integrating xSSC into leading baselines consistently establishes new SOTAs. Furthermore, our PCA and gradient analyses explicitly confirm the encoding of objects' time-invariant semantics and time-variant kinematics into the proposed sub-spaces.
\end{itemize}

\section{Related Work}
\label{sect:related_work}

\subsection{Video Object-Centric Learning}
\label{sect:vocl}

Video Object-Centric Learning (OCL) aims to discover objects without supervision by extending image-level OCL into temporal. Early plain video OCL methods, such as STEVE \cite{singh2022steve}, SAVi \cite{kipf2022savi} and SAVi++ \cite{elsayed2022savipp}, roughly follow a standard framework: using an auto-encoder structure where slots transition over time and aggregate visual features to reconstruct the input frames in some format \cite{zhao2025vvo}. To improve object decomposition and association, methods like SAVi and SAVi++ reconstruct optical flow and depth targets as weak supervision.

Recent advancements have largely focused on enhancing the transition and reconstruction modules to handle highly dynamic and real-world videos. For instance, VideoSAUR \cite{zadaianchuk2024videosaur} attempts to capture temporal dynamics by predicting patch movement in the next frame. More recently, a powerful lineage of SOTA architectures has emerged, relying on an explicit regularization introduced in SlotContrast \cite{manasyan2025slotcontrast}. It is an elegant Slot-Slot Contrastive (SSC) loss, which has since been inherited by subsequent strong baseline methods like RandSF.Q \cite{zhao2025randsfq} and SmoothSA \cite{zhao2025smoothsa}. While these contemporary methods achieve remarkable performance, they increasingly rely on adding auxiliary loss terms to the model.

In contrast, our work takes a step back. We demonstrate that by intelligently restructuring the slot representations themselves, the complex architectural bloat of recent SOTAs can be elegantly bypassed, adhering to Occam's Razor.

\subsection{Temporal Consistency}
\label{sect:tc}

Maintaining temporal consistency, i.e., the ability to track object identities and states across continuous frames, is one of the fundamental challenges in video OCL. Historically, researchers have tackled this using various explicit mechanisms, such as sequential VAEs \cite{yu2024vonet} or VQ-VAE \cite{zhao2025vvo}, agglomerative clustering \cite{aydemir2023solv}, or explicit memory buffers coupled with transformer predictors \cite{zhao2023ocmot}.

In the current landscape of Slot Attention-based \cite{locatello2020slotatt} video OCL, methods enforce temporal consistency primarily through explicit regularization. VideoSAUR \cite{zadaianchuk2024videosaur} proposes a time similarity loss. However, this explicitly assumes that object patches remain constantly visible in the subsequent frame, an assumption that severely degrades in highly dynamic datasets like MOVi-C/D/E \cite{greff2022kubric} where objects frequently occlude or disappear. To avoid such narrow assumptions, SlotContrast \cite{manasyan2025slotcontrast} and its successors, e.g., RandSF.Q \cite{zhao2025randsfq} and SmoothSA \cite{zhao2025smoothsa}, employ the InfoNCE-style \cite{oord2018infonce} SSC loss to explicitly pull slots that are in the same position of the same video sample close and push slots that are in different positions of the same or different samples apart across frames.

While explicitly regularizing slot disparities is effective, we argue it is fundamentally a workaround. Instead of enforcing temporal consistency explicitly outside, our method foster such property implicitly inside. By stochastically predicting cross-temporal features using a static-dynamic channel split, our method achieves the goal through the standard reconstruction objective alone.

\section{Proposed Method}
\label{sect:proposed_method}

Adhering to the elegance of Occam's Razor, we propose excluding the Slot-Slot Contrastive loss entirely (\textbf{xSSC}) from SOTA video OCL methods, as shown in \Cref{fig:teaser}. This enforces temporal consistency in slot sets implicitly through model design, rather than explicitly through extra losses.

\subsection{Preliminaries: Explicit Regularization Bottleneck}
\label{sect:preliminary}

Given a video, a Vision Foundation Model (VFM), like DINO-series \cite{caron2021dinov1, oquab2024dinov2}, encodes the $t$-th frame into features $\bm{F}_t \in \mathbb{R}^{h \times w \times c}$.

\textbf{A plain video OCL method} operates across all the frames of the video recurrently \cite{zhao2025vvo} as below:
\begin{subequations}
\begin{align}
\input{res/eq_vocl_zero}
\end{align}
\end{subequations}
Aggregator $\bm{\phi}_\mathrm{a}$, parameterized as a Slot Attention \cite{locatello2020slotatt} module, aggregates current features $\bm{F}_t$ into current slots $\bm{S}_t \in \mathbb{R} ^ {s \times c}$, with previous slots' transition as the query $\bm{Q}_t$.
The slot set $\bm{S}_t$ can be used to represent the visual scene as object-level feature vectors, while the corresponding attention maps $\bm{A}_t \in \mathbb{R} ^ {s \times h \times w}$ can be binarized as the byproduct segmentation masks of the objects and background.
Decoder $\bm{\phi}_\mathrm{d}$ decodes $\bm{S}_t$ into reconstructed current feature $\bm{F}'_t$.
The self-supervision signal comes from minimizing the reconstruction error $\mathcal{L}_\mathrm{recon}$.

To enforce objects' \textit{temporal consistency} across frames, recent SOTAs heavily rely on explicit losses, such as VideoSAUR's Temporal Similarity $\mathcal{L}_\mathrm{tsim}$ \cite{zadaianchuk2024videosaur} or SlotContrast's Slot-Slot Contrastive loss $\mathcal{L}_\mathrm{ssc}$ \cite{manasyan2025slotcontrast}.

\textbf{Temporal Similarity (TSim)}. VideoSAUR forces its decoder $\bm{\phi}^1_\mathrm{d}$ to additionally predict the future position $\bm{P}_{t+1} \in \mathbb{R} ^ {h \times w}$ of patches in current features $\bm{F}_t$ for extra supervision signal that promotes temporal consistency:
\begin{subequations}
\begin{align}
\input{res/eq_vocl_tsim}
\end{align}
\end{subequations}
where $\bm{P}'_{t+1} \in \mathbb{R} ^ {h \times w \times h' \times w'}$ is predicted future patch position, where $h'$ and $w'$ are identical to $h$ and $w$ respectively. Pseudo-ground-truth future patch position $\bm{P}_{t+1}$ is naively indicated by the maximum similarity along dimension $c$ between current features $\bm{F}_t$ and future features $\bm{F}_{t+1}$.

\textit{Remark}.
The additional prediction of $\bm{P}'_{t+1}$ consumes extra computation both in space and time, so does the corresponding loss minimizing.
The calculation of pseudo-ground-truth $\bm{P}_{t+1}$ actually assumes that patches in current frame are still visible in next frame. Thus minimizing $\mathcal{L}_\mathrm{tsim}$ can be harmful for learning on videos where objects have strong dynamic occlusion, disappearing and re-appearing, e.g., MOVi-series datasets \cite{greff2022kubric}. Please refer to \Cref{tab:objdiscov} -- VideoSAUR shows a significant gap from SlotContrast on especially MOVi-C/E.

\textbf{Slot-Slot Contrastive loss (SSC)}. SlotContrast adopts the plain model design and introduces an elegant contrastive loss that pull slots representing the same object close while pushing slots representing different object apart, based on an InfoNCE-style \cite{oord2018infonce} regularization:
\begin{subequations}
\begin{align}
\input{res/eq_vocl_ssc}
\end{align}
\end{subequations}
where $\{ \bm{S}_t^{(b)} \}_b \in \mathbb{R} ^ {b \times s \times c}$ is $b$ slot sets in a batch, as the contrastive loss is calculated along both dimensions $b$ and $s$.

\textit{Remark}.
Compared with TSim, SSC is elegant both in model design and loss minimizing. It works effectively on different datasets and thus is inherited by the most recent strong SOTAs like RandSF.Q and SmoothSA.
However, besides also introducing extra computation overhead, it relies on stable one-to-one object correspondence across adjacent frames.

\textbf{Our xSSC shifts the paradigm} for learning temporal consistency from explicit loss terms to implicit model designs. It avoids computation overhead in both training and inference, as well as corresponding hyperparamters. It also replies on a minimal assumption that the current frame's whole set of slots contains all the static information about the previous frame.

\subsection{Chrono-Channel Decomposition (CCD)}
\label{sect:ccd}

The fundamental philosophy of xSSC is that an object's temporal presence consists of two physically distinct components: its persistent/\textit{static} appearance and its changing/\textit{dynamic} kinematics. The former component is used to foster temporal consistency while the latter preserves representation diversity.

We formulate this by introducing Chrono-Channel Decomposition (CCD).
Instead of treating each slot in the set $\bm{S}_t$ as a monolithic latent vector, we structurally disentangle it along the channel dimension into a static sub-space and a dynamic sub-space.
Let $c_\mathrm{s}$ denote the channel boundary. The static part $\bm{S}_{t, :c_\mathrm{s}}$ encodes time-invariant object identity like semantics and texture, while the dynamic part $\bm{S}_{t, c_\mathrm{s}:}$ encodes time-variant states like position, pose and motion.

Crucially, CCD acts as an implicit information bottleneck. Since kinematic states operate on a significantly lower-dimensional manifold compared to rich visual semantics, we strictly constrain the dynamic channels (empirically, one-quarter of the total channels, i.e., $c_\mathrm{s}=\frac{3}{4}c$). This physical constraint prevents the network from redundantly memorizing visual features in the dynamic sub-space, forcing a clean disentanglement. Please refer to \Cref{sect:dissect}.

\subsection{Cross-Temporal Reconstruction (CTR)}
\label{sect:ctr}

To enforce the structural disentanglement of CCD without any explicit contrastive penalties, we introduce a parameter-free training mechanism: Cross-Temporal Reconstruction (CTR).

During training, instead of conventionally reconstructing current frame $\bm{F}_t$, a typical decoder $\bm{\phi}_\mathrm{d}$ is tasked to stochastically reconstruct target features of either the current time step ($t_x = t$) or the previous ($t_x = t-1$). To achieve this, we combine the current frame's static properties with the target frame's dynamic states.
Formally, the decoder $\bm{\phi}_\mathrm{d}$ operates on synthesized slot $\tilde{\bm{S}}_t \in \mathbb{R} ^ {s \times c}$:
\begin{subequations}
\begin{align}
\input{res/eq_vocl_xssc}
\end{align}
\end{subequations}
where $\Tilde{\bm{S}}_t$ are constructed via cross-temporal channel concatenation along with relative temporal embeddings $\bm{E} \in \mathbb{R} ^ {2 \times c}$.
$\oplus$ means appending relative time embeddings to slots.
During inference, $t_x \equiv t$, working as common decoding.

\subsection{Implicit Temporal Consistency via Standard Reconstruction}
\label{sect:objective}

Through the synergy of CCD and CTR, xSSC perfectly internalizes temporal consistency. The final objective is super simple, returning to the purest form:
\begin{equation}
\label{eq:xssc_loss}
\input{res/eq_vocl_xssc_loss}
\end{equation}

\textit{Remark}.
By completely dropping explicit losses like $\mathcal{L}_\mathrm{tsim}$ $\mathcal{L}_\mathrm{ssc}$, we eliminate hyper-parameter tuning.
Furthermore, CCD's naive projection and channel group/concatenation is very light-wegiht; CTR utilizes a single standard OCL decoding head to produce the same volume of reconstruction as the plain decoding. Thus during both training and inference, quasi-zero computation overhead is introduced neither in space nor in time.

\begin{figure}
\centering
\includegraphics[width=\linewidth]{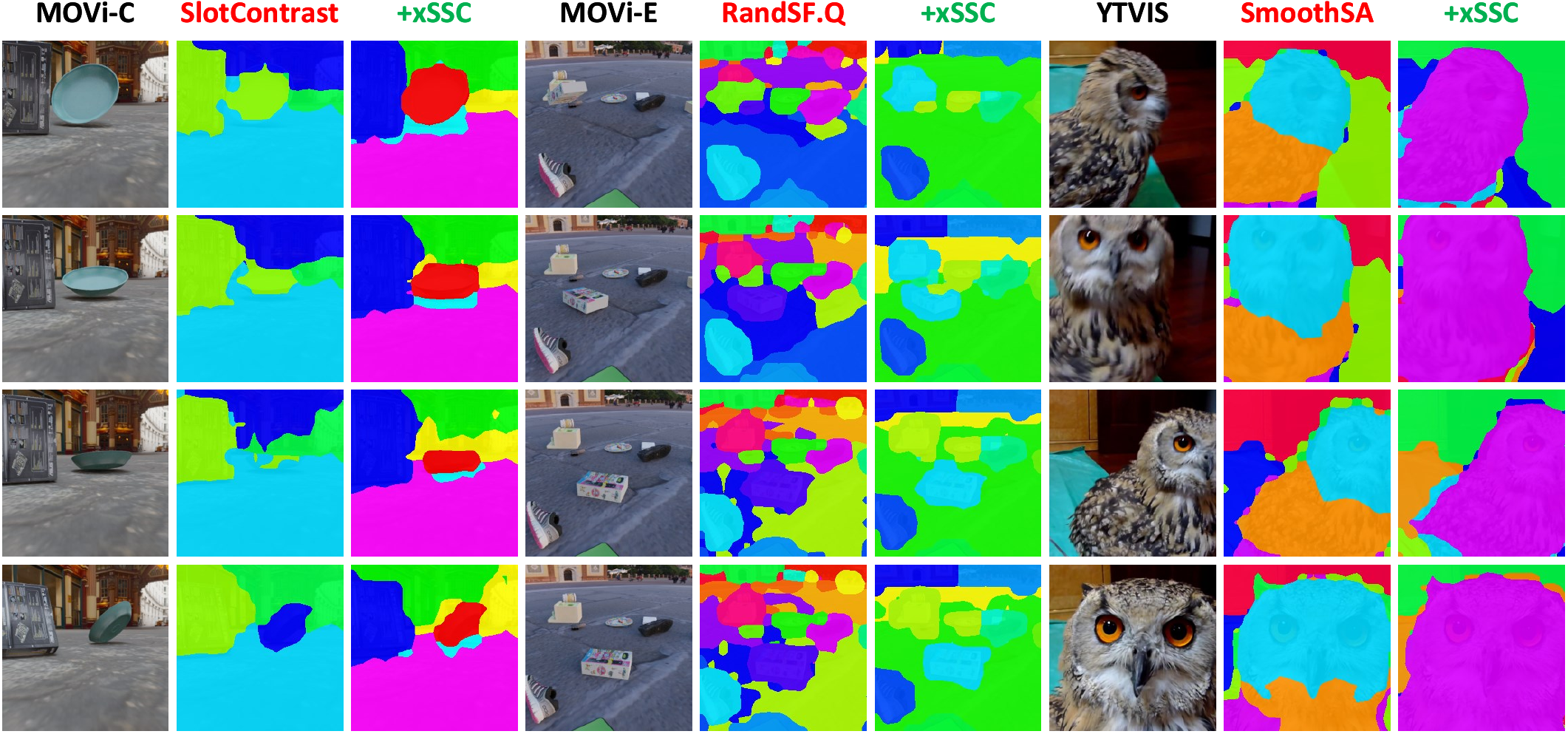}
\caption{Object Discovery Visualization.}
\label{fig:qualitative}
\end{figure}

\section{Experiment}
\label{sect:experiment}

We extensively evaluate xSSC across unsupervised object discovery (\Cref{sect:objdiscov}), downstream object recognition (\Cref{sect:objrecogn}), and rigorously dissect its internal mechanisms through attribution and manipulation analyses (\Cref{sect:dissect}). To ensure absolute reproducibility, we use three standard random seeds, i.e., 42, 43 and 44, for any experiment item whenever possible.

\subsection{Experiment Setup and Unified Benchmark}
\label{sect:experiment_setup}

\textbf{Datasets}. We evaluate on datasets exhibiting varying degrees of visual and temporal complexity. MOVi-C and MOVi-E~\footnote{https://github.com/google-research/kubric/blob/main/challenges/movi} are synthetic datasets featuring complex object dynamics, with MOVi-E further introducing camera movements. YTVIS~\footnote{https://youtube-vos.org/dataset/vis} represents real-world YouTube videos characterized by highly complex backgrounds, frequent occlusions, and diverse object categories. We use the High-Quality (HQ) version~\footnote{https://github.com/SysCV/vmt?tab=readme-ov-file\#hq-ytvis-high-quality-video-instance-segmentation-dataset}, which provides better segmentation annotations.

\textbf{Metrics}. For \textit{object discovery}, we utilize the standard segmentation metrics: Adjusted Rand Index (ARI)~\footnote{https://scikit-learn.org/stable/modules/generated/sklearn.metrics.adjusted\_rand\_\\score.html}, Foreground ARI (ARI\textsubscript{fg}), Mean Best Overlap (mBO) \cite{uijlings2013selectivesearch}, and Mean Intersection over Union (mIoU)~\footnote{https://scikit-learn.org/stable/modules/generated/sklearn.metrics.jaccard\_score.ht\\ml}.
For downstream \textit{object recognition}, we report Top-1 and Top-3 classification accuracy, bounding box IoU (box IoU), and the number of matched objects (\#match).

\textbf{Codebase and Fair Comparison}. A critical issue in video OCL literature is that methods often report results using disjoint codebases, with different data augmentations, training schedules, input resolutions, VFM backbone sizes and so on.
To ensure a strictly controlled and fair comparison, we conduct all experiments within a unified SOTA framework \texttt{object-centric-bench}~\footnote{https://github.com/Genera1Z/RandSF.Q}~\footnote{https://github.com/Genera1Z/SmoothSA}. Many representative methods are reproduced with identical, advanced data augmentation and training recipes, even with model checkpoints and training logs for all random seeds available. Consequently, any performance delta in our experiments can be attributed to the claimed differences, rather than hyper-parameters.

\begin{table}[]
\centering
\input{res/tab_objdiscov}
\caption{Object Discovery. For all methods, we use DINOv2 ViT-S/14 \cite{oquab2024dinov2} as the encoder, strong data augmentation \cite{elsayed2022savipp}, input solution 224$\times$224.}
\label{tab:objdiscov}
\end{table}

\subsection{Unsupervised Video Object Discovery}
\label{sect:objdiscov}

\textbf{Setup}. We seamlessly integrate our xSSC into three leading baselines, SlotContrast \cite{manasyan2025slotcontrast}, RandSF.Q \cite{zhao2025randsfq} and SmoothSA \cite{zhao2025smoothsa}, strictly removing their explicit SSC losses.
We also report results of VideoSAUR \cite{zadaianchuk2024videosaur} as a reference point.
We skip methods like SAVi \cite{kipf2022savi} and SAVi++ \cite{elsayed2022savipp}, whose extra weak supervision based optical flow and depth is not fair for our baselines.
We also skip methods like SOLV \cite{aydemir2023solv} and DIAS \cite{zhao2025dias}, whose slot pruning/merging technique is not fair for our baselines, which use fixed number of slots.

\textbf{Results}. As shown in \Cref{tab:objdiscov}, replacing explicit contrastive losses with our implicit xSSC consistently establishes new SOTAs across all baselines and datasets under most metrics. Crucially, in highly dynamic scenes (MOVi-C/E) where objects frequently occlude or reappear, the rigid tracking assumptions of SSC' explicit regularization become bottlenecks. By relaxing this constraint and aligning identities implicitly through the static sub-space, SlotContrast+xSSC, RandSF.Q+xSSC and SmoothSA+xSSC demonstrate consistent gains in all metrics, especially ARI, mBO and mIoU.
Qualitative results are shown in \Cref{fig:qualitative}.
Furthermore, as shown in \Cref{tab:objdiscov_efficiency}, by excluding the dense explicit contrastive computation, xSSC practically improves training efficiency in both memory and throughput.
This proves that Occam's Razor holds: simpler, structural designs outperform complex loss engineering.

\begin{table}
\begin{minipage}{0.48\textwidth}
\centering
\input{res/tab_objdiscov_efficiency}
\caption{Training/Evalution Efficiency.}
\label{tab:objdiscov_efficiency}
\end{minipage}
\begin{minipage}{0.5\textwidth}
\centering
\input{res/tab_objrecogn}
\caption{Object Recognition.}
\label{tab:objrecogn}
\end{minipage}
\end{table}

\subsection{Supervised Probing: Object Recognition}
\label{sect:objrecogn}

\textbf{Setup}. Object discovery merely measures spatial decomposition plus temporal association, while object recognition explicitly quantifies the semantic quality of the learned latent slots. 
Following standard protocols \cite{seitzer2023dinosaur}, we freeze the trained OCL models from \Cref{sect:objdiscov} and train a lightweight two-layer MLP to predict class labels and bounding boxes via the classification and regression heads respectively, from the slot representations on the YTVIS-HQ dataset.

\textbf{Results}. As shown in \Cref{tab:objrecogn}, models powered by xSSC consistently outperform their vanilla counterparts in Top-1/Top-3 accuracy and box IoU, under similar number of matched objects \#match. We argue that the explicit SSC loss often forces slots to over-fit to instance-level low-level textures to satisfy the InfoNCE penalty. In contrast, xSSC explicitly allocates a static channel sub-space for identity, preventing dynamic noises (pose, location) from diluting the semantic features. This structural bottleneck ensures that the frozen slots provide a far more linearly separable latent space for downstream reasoning.

\begin{figure}
\centering
\includegraphics[width=0.41\linewidth]{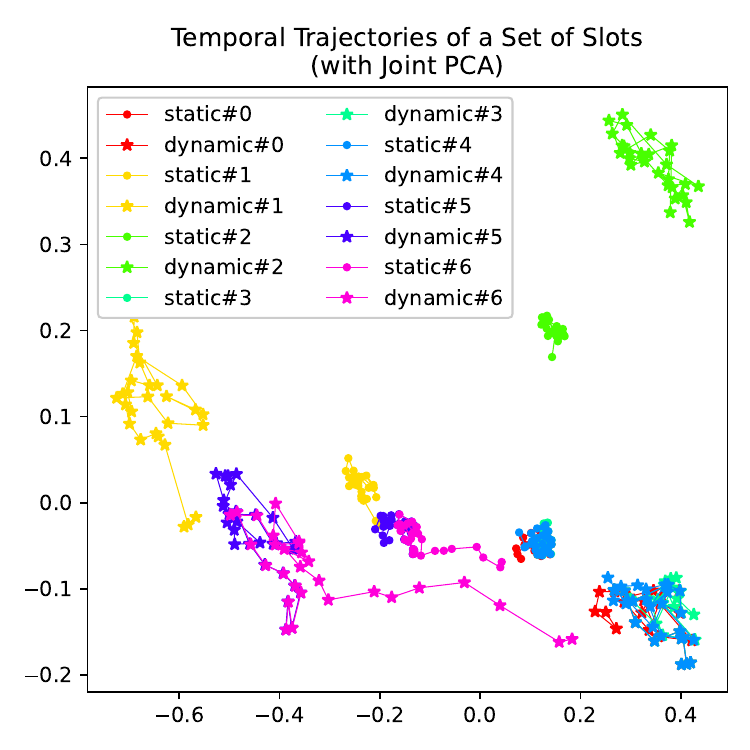}
\includegraphics[width=0.1625\linewidth]{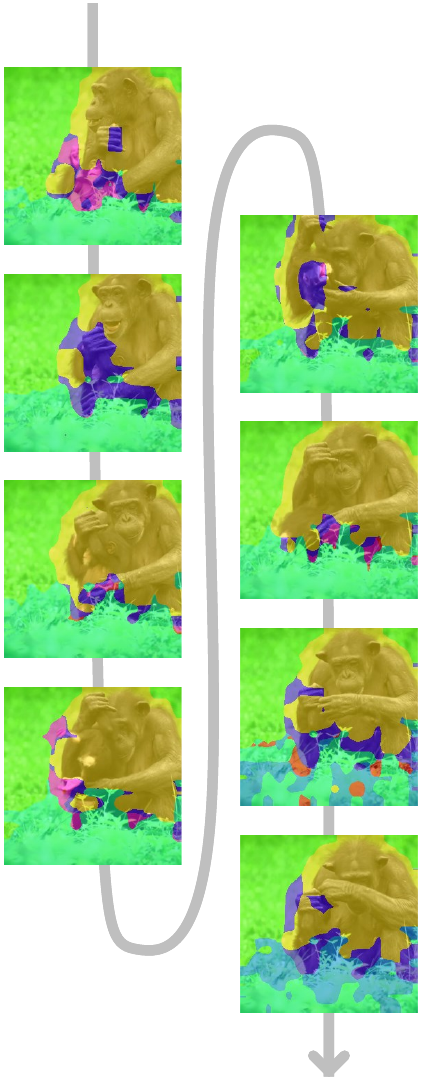}
\includegraphics[width=0.41\linewidth]{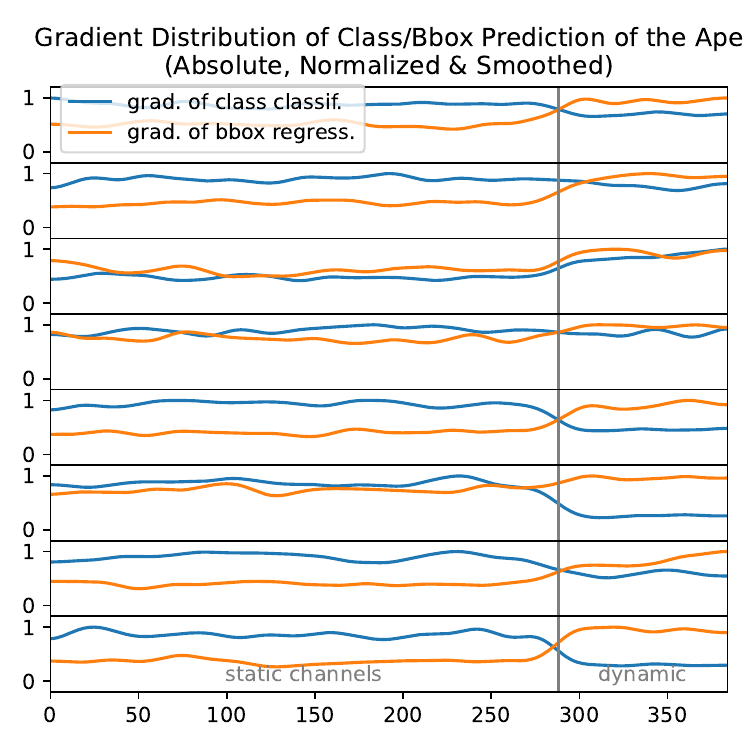}
\caption{
(\textit{center}) A sample video with object discovery masks, 8 of 29 frames.
(\textit{left}) Temporal Trajectories: slots' static part usually clusters while the dynamic part mostly disperses. Interestingly, the purple slot's static part also spreads, due to its noisy feature aggregation, which jumps across frames.
(\textit{right}) Gradient Distribution: to classify the slot that aggregates the yellow area features, more gradients come from the static channels, which verifies our design that static channels capture more invariant semantics; Similarly, to regress the ape's bounding box, more kinematics is needed.
}
\label{fig:dissect}
\end{figure}

\subsection{Dissection: Static vs Dynamic Sub-spaces}
\label{sect:dissect}

To rigorously validate that with the help of CTR, CCD physically disentangles semantics and kinematics as claimed, we conduct three white-box analyses on the converged SmoothSA+xSSC model. The corresponding results of the SmoothSA basis model is included as a reference.

\textbf{Temporal PCA Trajectories}.
As shown in \Cref{fig:dissect} (\textit{left}), we project the slots of a tracked object across a video into a 2D space using PCA. When projecting only the static channels, i.e., $\{ \bm{S}_{t, :c_\mathrm{s}} \}_t$, the trajectory collapses into a tightly bounded dense cluster. This confirms that time-invariant identity is successfully anchored. Conversely, projecting the dynamic channels, i.e., $\{ \bm{S}_{t, c_\mathrm{s}:} \}_t$, reveals a continuous, elongated trajectory. It effectively and intuitively maps the object's spatial movement and kinematic state changes over time.

\textbf{Gradient Attribution in Object Recognition}.
As shown in \Cref{fig:dissect} (\textit{right}), we compute the gradient magnitude from the downstream MLP back to the frozen slots. For the categorical classification head (relying on identity), most of the gradient attribution is heavily concentrated on the static channels. For the bounding box regression head (relying on spatial coordinates), the gradients distinctly shift, primarily activating the dynamic channels. This proves our proposed sub-spaces function exactly as an information bottleneck.

\subsection{Ablation}
\label{sect:ablat}

Our ablation study is shown in \Cref{tab:ablat}.

\textbf{Effects of relative time embedding}: with this is always better. Thus it is necessary to indicate the decoder what are the target to predict.

\textbf{Effects of dynamic channel ratio}: A quarter is the best choice, compared with $\frac{1}{2}$ and $\frac{1}{8}$. It performs consistently well on quite different datasets.

\textbf{Effects of reconstruction past or future}: Reconstruction the past is better than the future. This is because the future slots are recurrently transitioned from the past thus they should contain the information about the past, which aligns with our assumption.

\textbf{Effects of probability of predicting the past}: 50/50 is the best choice as it balances the learning of standard reconstruction and the ability to remember the past.

\begin{table}[]
\centering
\input{res/tab_ablat}
\caption{Ablation Studies.}
\label{tab:ablat}
\end{table}

\section{Limitation and Future Work}
\label{sect:discuss}

While xSSC establishes a new paradigm for video object-centric learning by internalizing temporal consistency via structural design, it naturally inherits certain limitations from its purely reconstruction-driven nature.

Firstly, relying solely on the plain reconstruction loss without explicit instance-level push-away mechanics, which is intrinsically balanced in InfoNCE by the pull-close, the model may struggle in extremely ambiguous scenes featuring multiple visually identical instances. For example, given a swarm of identical uniform objects, the static sub-spaces might overly converge, burdening the dynamic channels to separate them spatially.

Secondly, our static-dynamic decomposition implicitly assumes that an object's identity lies in low-frequency static semantics. For things or stuff with highly dynamic intrinsic textures, e.g., flickering fire and rippling water, the compact dynamic sub-space may face bandwidth bottlenecks, leading to blurry reconstructions.

Finally, addressing long-term re-identification, where objects are occluded for extended periods beyond the stochastic adjacent-frame window, remains an open challenge. Future work could explore integrating slot-based memory buffers into the static sub-space to anchor persistent identities across long-term occlusions, yet without sacrificing Occam's Razor.

\section{Conclusion}
\label{sect:conclusion}

In this work, we propose a paradigm shift for video Object-Centric Learning: robust temporal consistency is fundamentally better enforced as an implicit architectural property rather than an explicit external penalty. We demonstrate that the explicit SSC regularization and its rigid tracking assumptions, which have dominated recent state-of-the-art methods, can be elegantly and completely excluded.
We achieve this via two quasi-zero-overhead synergistic mechanisms, Chrono-Channel Decomposition (CCD) and Cross-Temporal Reconstruction (CTR). Together, they compel the model to naturally internalize temporal correspondence by minimizing the standard reconstruction error alone.
Extensive empirical evaluations and rigorous white-box analyses confirm that xSSC not only establishes new SOTA performance on highly dynamic video datasets but also yields profoundly interpretable latent sub-spaces.
While resolving extremely ambiguous identical instances and long-term occlusions purely through reconstruction remains an open challenge, xSSC strictly adheres to Occam’s Razor. We successfully return video OCL to its purest form, providing high efficiency and elegant foundation for future research.


%
%
\bibliographystyle{splncs04}
\bibliography{main}

\end{document}

%% file: res/eq_vocl_zero.tex
\label{eq:aggregat}
\bm{S}_t, \bm{A}_t &= \bm{\phi}_\text{a} ( \bm{Q}_t , \bm{F}_t )
\\
\label{eq:decode}
\bm{F}'_t &= \bm{\phi}_\text{d} ( \bm{S}_t )
\\
\label{eq:loss}
\mathcal{L} & = \mathcal{L}_\mathrm{recon} = \mathrm{MSE} ( {\bm{F}'_t} , \bm{F}_t )

%% file: res/eq_vocl_tsim.tex
\label{eq:decode}
&& \bm{F}'_t , \bm{P}'_{t+1} &= \bm{\phi}^1_\text{d} ( \bm{S}_t )
\\
\label{eq:loss}
&& \mathcal{L} &= \mathcal{L}_\mathrm{recon} + \lambda \mathcal{L}_\mathrm{tsim}
\\
\label{eq:tsim}
& \text{with}
& \mathcal{L}_\mathrm{tsim} &= \mathrm{mean} _ {h,w} ( \mathrm{CE} ( \bm{P}'_{t+1} , \bm{P}_{t+1} ) )
\\
\label{eq:pgt}
& \text{and}
& \bm{P}_{t+1} &= \mathrm{arg max} _ {h',w'} ( \mathrm{similarity}_c ( \bm{F}_t , \bm{F}_{t+1} ) )

%% file: res/eq_vocl_ssc.tex
\label{eq:loss}
& & \mathcal{L} &= \mathcal{L}_\mathrm{recon} + \lambda \mathcal{L}_\mathrm{ssc}
\\
\label{eq:ssc}
& \text{with}
& \mathcal{L}_\mathrm{ssc} &= \mathrm{InfoNCE}_{b,s} ( \{ \bm{S}^{(b)}_{t} \}_b , \{ \bm{S}^{(b)}_{t+1} \}_b )

%% file: res/eq_vocl_xssc.tex
\label{eq:decode}
& & \bm{F}'_{t_x} &= \bm{\phi}_\text{d} ( \Tilde{ \bm{S}_t } )
\\
& \text{with}
& \Tilde{ \bm{S}_t } &= \mathrm{cat}_c ( \bm{S}_{t,:c_\text{s}} , \bm{S}_{t_x,c_\text{s}:} ) \oplus \mathrm{cat}_c ( \bm{E}_{t-t_x,:c_\text{s}} , \bm{E}_{0,c_\text{s}:} )

%% file: res/eq_vocl_xssc_loss.tex
\mathcal{L} = \mathcal{L}_\mathrm{recon} = \mathrm{MSE} (\bm{F}'_{t_x} , \bm{F}_{t_x})

%% file: res/tab_objdiscov.tex
\setlength{\tabcolsep}{1.125pt}
\newcommand{\tss}[1]{\scalebox{0.4}{#1}}
\newcommand{\std}[1]{\scalebox{0.4}{±#1}}
\newcommand{\cg}[1]{\textcolor{green}{#1}}
\newcommand{\ch}[1]{\textcolor{red}{#1}}

\begin{tabular}{ccccccccccccc}
\toprule
& ARI & ARI\textsubscript{fg} & mBO & mIoU
& ARI & ARI\textsubscript{fg} & mBO & mIoU
& ARI & ARI\textsubscript{fg} & mBO & mIoU
\\
\arrayrulecolor{gray}
\cmidrule(lr){2-5}
\cmidrule(lr){6-9}
\cmidrule(lr){10-13}
\arrayrulecolor{black}
& \multicolumn{4}{c}{MOVi-C \tss{\#slot=11, conditional}}
& \multicolumn{4}{c}{MOVi-E \tss{\#slot=24, conditional}}
& \multicolumn{4}{c}{YTVIS-HQ \tss{\#slot=6, unconditional}}
\\
\cmidrule(lr){1-1}
\cmidrule(lr){2-5}
\cmidrule(lr){6-9}
\cmidrule(lr){10-13}
VideoSAUR
& 41.9\tss{1.1} & 53.3\tss{2.1} & 16.1\tss{0.4} & 14.8\tss{0.4}
& 17.4\tss{2.5} & 34.6\tss{20.7} & 8.3\tss{4.9} & 7.5\tss{4.3}
& 33.8\tss{0.7} & 49.2\tss{0.5} & 29.9\tss{0.4} & 29.7\tss{0.4}
\\
\arrayrulecolor{gray}
\cmidrule(lr){1-1}
\cmidrule(lr){2-5}
\cmidrule(lr){6-9}
\cmidrule(lr){10-13}
SlotContrast
& 64.6\tss{9.4} & 59.9\tss{5.3} & 27.7\tss{3.0} & 25.8\tss{2.9}
& 29.9\tss{4.9} & 70.6\tss{3.8} & 20.7\tss{1.4} & 19.3\tss{1.2}
& 37.2\tss{0.6} & 49.4\tss{1.1} & 33.0\tss{0.2} & 32.8\tss{0.1}
\\
+\cg{xSSC}
& \cg{65.8}\tss{7.8} & \cg{61.5}\tss{4.9} & \cg{28.3}\tss{3.3} & \cg{25.9}\tss{2.8}
& \cg{34.0}\tss{3.7} & \cg{68.0}\tss{2.9} & \cg{21.9}\tss{2.0} & \cg{20.8}\tss{2.2}
& \cg{39.3}\tss{1.6} & \cg{52.5}\tss{1.8} & \cg{35.1}\tss{2.6} & \cg{34.3}\tss{1.0}
\\
\cmidrule(lr){1-1}
\cmidrule(lr){2-5}
\cmidrule(lr){6-9}
\cmidrule(lr){10-13}
RandSF.Q
& 65.4\tss{10.7} & 67.4\tss{2.1} & 29.2\tss{3.8} & 26.8\tss{3.7}
& 30.5\tss{1.2} & 82.1\tss{3.1} & 23.0\tss{1.2} & 21.6\tss{1.4}
& 40.1\tss{0.4} & 58.0\tss{1.0} & 37.6\tss{0.4} & 37.2\tss{0.4}
\\
+\cg{xSSC}
& \cg{68.4}\tss{5.4} & \cg{68.4}\tss{1.5} & \cg{30.8}\tss{2.4} & \cg{28.5}\tss{2.0}
& \cg{43.9}\tss{0.6} & \ch{75.2}\tss{1.2} & \cg{26.7}\tss{0.3} & \cg{25.0}\tss{0.3}
& \cg{41.8}\tss{0.9} & \cg{58.2}\tss{1.3} & \cg{39.5}\tss{0.5} & \cg{39.2}\tss{0.5}
\\
\cmidrule(lr){1-1}
\cmidrule(lr){2-5}
\cmidrule(lr){6-9}
\cmidrule(lr){10-13}
SmoothSA
& 50.9\tss{1.6} & 69.0\tss{0.3} & 31.7\tss{0.8} & 30.2\tss{0.8}
& 36.7\tss{0.6} & 73.6\tss{0.6} & 28.6\tss{0.1} & 27.4\tss{0.1}
& 42.4\tss{0.8} & 63.0\tss{3.4} & 38.9\tss{0.7} & 38.3\tss{0.6}
\\
+\cg{xSSC}
& \cg{55.5}\tss{0.3} & \cg{72.2}\tss{0.4} & \cg{34.5}\tss{0.3} & \cg{33.3}\tss{0.2}
& \ch{36.3}\tss{1.4} & \cg{74.5}\tss{1.3} & \cg{28.9}\tss{0.7} & \cg{27.7}\tss{0.7}
& \cg{45.5}\tss{1.4} & \ch{61.4}\tss{3.0} & \cg{41.1}\tss{0.4} & \cg{40.4}\tss{0.4}
\\
\arrayrulecolor{black}
\bottomrule
\end{tabular}

%% file: res/tab_objdiscov_efficiency.tex
\newcommand{\tss}[1]{\scalebox{0.4}{#1}}
\newcommand{\std}[1]{\scalebox{0.4}{±#1}}
\newcommand{\cg}[1]{\textcolor{green}{#1}}
\newcommand{\ch}[1]{\textcolor{red}{#1}}

\begin{tabular}{ccccc}
\toprule
& \multicolumn{2}{c}{train GB/min} & \multicolumn{2}{c}{eval}
\\
\arrayrulecolor{gray}
\cmidrule(lr){2-3} \cmidrule(lr){4-5}
& \multicolumn{4}{c}{YTVIS-HQ \tss{batchsize=8, RTX 3080, one epoch}}
\\
\arrayrulecolor{black}
\cmidrule(lr){1-1}
\cmidrule(lr){2-5}
\arrayrulecolor{gray}
SlotContrast
& 4.63 & 1.20 & 1.74 & 0.43
\\
+\cg{xSSC}
& \cg{4.46} & \cg{1.10} & \cg{1.66} & \cg{0.42}
\\
\cmidrule(lr){1-1}
\cmidrule(lr){2-3} \cmidrule(lr){4-5}
RandSF.Q
& 3.93 & 1.13 & 1.04 & 0.37
\\
+\cg{xSSC}
& \cg{3.82} & \cg{1.12} & \cg{1.03} & 0.37
\\
\cmidrule(lr){1-1}
\cmidrule(lr){2-3} \cmidrule(lr){4-5}
SmoothSA
& 3.97 & 1.08 & 1.04 & 0.40
\\
+\cg{xSSC}
& \cg{3.95} & \ch{1.10} & \cg{1.03} & 0.40
\\
\arrayrulecolor{black}
\bottomrule
\end{tabular}

%% file: res/tab_objrecogn.tex
\newcommand{\tss}[1]{\scalebox{0.4}{#1}}
\newcommand{\std}[1]{\scalebox{0.4}{±#1}}
\newcommand{\cg}[1]{\textcolor{green}{#1}}
\newcommand{\ch}[1]{\textcolor{red}{#1}}

\begin{tabular}{ccccc}
\toprule
& \multicolumn{2}{c}{cls Top-1/3} & box IoU & \#match
\\
\arrayrulecolor{gray}
\cmidrule(lr){2-3} \cmidrule(lr){4-4} \cmidrule(lr){5-5}
& \multicolumn{4}{c}{YTVIS-HQ \tss{freeze OCL + train MLP}}
\\
\arrayrulecolor{black}
\cmidrule(lr){1-1}
\cmidrule(lr){2-5}
\arrayrulecolor{gray}
SlotContrast
& 85.8\tss{0.3} & 95.8\tss{0.4} & 51.5\tss{0.5} & 9249\tss{41}
\\
+\cg{xSSC}
& \cg{86.4}\tss{1.2} & \ch{95.4}\tss{1.5} & \cg{52.4}\tss{0.6} & \cg{9253}\tss{58}
\\
\cmidrule(lr){1-1}
\cmidrule(lr){2-3} \cmidrule(lr){4-4} \cmidrule(lr){5-5}
RandSF.Q
& 90.5\tss{0.3} & 97.9\tss{0.3} & 50.6\tss{0.4} & 8979\tss{123}
\\
+\cg{xSSC}
& \cg{91.9}\tss{0.5} & \cg{97.9}\tss{0.1} & \cg{52.9}\tss{0.7} & \cg{9183}\tss{34}
\\
\cmidrule(lr){1-1}
\cmidrule(lr){2-3} \cmidrule(lr){4-4} \cmidrule(lr){5-5}
SmoothSA
& 90.4\tss{0.2} & 97.6\tss{0.1} & 42.6\tss{1.4} & 8957\tss{34}
\\
+\cg{xSSC}
& \cg{91.4}\tss{0.5} & \cg{98.2}\tss{0.0} & \cg{43.5}\tss{0.3} & \ch{8777}\tss{89}
\\
\arrayrulecolor{black}
\bottomrule
\end{tabular}

%% file: res/tab_ablat.tex
\setlength{\tabcolsep}{1pt}
\newcommand{\tss}[1]{\scalebox{0.4}{#1}}
\newcommand{\std}[1]{\scalebox{0.4}{±#1}}
\newcommand{\cg}[1]{\textcolor{green}{#1}}
\newcommand{\ch}[1]{\textcolor{red}{#1}}

\begin{tabular}{cccccccccccc}
\toprule
ARI+ARI\textsubscript{fg}+mBO+mIoU
& \multicolumn{2}{c}{relative t emb.} & \multicolumn{3}{c}{dynamic ratio} 
& \multicolumn{2}{c}{recon.}
& \multicolumn{3}{c}{$\mathrm{P}(\text{past})$}
\\
\arrayrulecolor{gray}
\cmidrule(lr){2-3} \cmidrule(lr){4-6} \cmidrule(lr){7-8} \cmidrule(lr){9-11}
YTVIS-HQ
& \cg{with} & without & $\frac{1}{2}$ & \cg{$\frac{1}{4}$} & $\frac{1}{8}$ 
& \cg{past} & future
& 0.1 & \cg{0.5} & 0.9
\\
\arrayrulecolor{black}
\cmidrule(lr){1-1}
\cmidrule(lr){2-3} \cmidrule(lr){4-6} \cmidrule(lr){7-8} \cmidrule(lr){9-11}
\arrayrulecolor{gray}
RandSF.Q+xSSC
& \cg{178.7}\tss{3.0} & 164.8\tss{9.7}
& 176.3\tss{6.8} & \cg{-} & 178.6\tss{6.2}
& \cg{-} & 174.2\tss{7.4}
& 175.2\tss{9.3} & \cg{-} & 170.6\tss{2.5} 
\\
\cmidrule(lr){1-1}
\cmidrule(lr){2-3} \cmidrule(lr){4-6} \cmidrule(lr){7-8} \cmidrule(lr){9-11}
SmoothSA+xSSC
& \cg{188.4}\tss{4.7} & 169.3\tss{8.3}
& 185.4\tss{8.2} & \cg{-} & 187.6\tss{9.3}
& \cg{-} & 173.7\tss{8.2}
& 184.6\tss{4.4} & \cg{-} & 179.9\tss{6.5}
\\
\arrayrulecolor{black}
\bottomrule
\end{tabular}

%% file: main.bib
@String(NeurIPS = {Adv. Neural Inform. Process. Syst.})

@String(ICLR  = {Int. Conf. Learn. Represent.})

@String(AAAI  = {AAAI})

@String(NeurIPS = {NeurIPS})

@String(ICLR  = {ICLR})

@inproceedings{locatello2020slotatt,
 author = {Locatello, Francesco and Weissenborn, Dirk and Unterthiner, Thomas and Mahendran, Aravindh and Heigold, Georg and Uszkoreit, Jakob and Dosovitskiy, Alexey and Kipf, Thomas},
 booktitle = {Advances in Neural Information Processing Systems},
 editor = {H. Larochelle and M. Ranzato and R. Hadsell and M.F. Balcan and H. Lin},
 pages = {11525--11538},
 publisher = {Curran Associates, Inc.},
 title = {{Object-Centric Learning with Slot Attention}},
 volume = {33},
 year = {2020}
}

@article{singh2022steve,
  title={{Simple Unsupervised Object-Centric Learning for Complex and Naturalistic Videos}},
  author={Singh, Gautam and Wu, Yi-Fu and Ahn, Sungjin},
  journal={Advances in Neural Information Processing Systems},
  volume={35},
  pages={18181--18196},
  year={2022}
}

@inproceedings{kipf2022savi,
    author = {Kipf, Thomas and Elsayed, Gamaleldin F. and Mahendran, Aravindh
              and Stone, Austin and Sabour, Sara and Heigold, Georg
              and Jonschkowski, Rico and Dosovitskiy, Alexey and Greff, Klaus},
    title = {{Conditional Object-Centric Learning from Video}},
    booktitle = {International Conference on Learning Representations (ICLR)},
    year  = {2022}
}

@inproceedings{elsayed2022savipp,
    author={Elsayed, Gamaleldin F. and Mahendran, Aravindh
    and van Steenkiste, Sjoerd and Greff, Klaus and Mozer, Michael C.
    and Kipf, Thomas},
    title = {{SAVi++: Towards End-to-End Object-Centric Learning from Real-World Videos}},
    booktitle = {Advances in Neural Information Processing Systems (NeurIPS)},
    year  = {2022}
}

@article{zadaianchuk2024videosaur,
  title={{Object-Centric Learning for Real-World Videos by Predicting Temporal Feature Similarities}},
  author={Zadaianchuk, Andrii and Seitzer, Maximilian and Martius, Georg},
  journal={Advances in Neural Information Processing Systems},
  volume={36},
  year={2024}
}

@inproceedings{manasyan2025slotcontrast,
  title={{Temporally Consistent Object-Centric Learning by Contrasting Slots}},
  author={Manasyan, Anna and Seitzer, Maximilian and Radovic, Filip and Martius, Georg and Zadaianchuk, Andrii},
  booktitle={Proceedings of the Computer Vision and Pattern Recognition Conference},
  pages={5401--5411},
  year={2025}
}

@inproceedings{zhao2025vvo,
  title={{Vector-Quantized Vision Foundation Model for Object-Centric Learning}},
  author={Zhao, Rongzhen and Wang, Vivienne and Kannala, Juho and Pajarinen, Joni},
  booktitle={ACM Multimedia},
  year={2025}
}

@inproceedings{zhao2025randsfq,
  title={{Predicting Video Slot Attention Queries from Random Slot-Feature Pairs}},
  author={Zhao, Rongzhen and Li, Jian and Kannala, Juho and Pajarinen, Joni},
  booktitle={AAAI},
  year={2026}
}

@article{zhao2025smoothsa,
  title={{Smoothing Slot Attention Iterations and Recurrences}},
  author={Zhao, Rongzhen and Yang, Wenyan and Kannala, Juho and Pajarinen, Joni},
  journal={arXiv preprint arXiv:2508.05417},
  year={2025}
}

@article{oord2018infonce,
  title={{Representation Learning with Contrastive Predictive Coding}},
  author={Oord, Aaron van den and Li, Yazhe and Vinyals, Oriol},
  journal={arXiv preprint arXiv:1807.03748},
  year={2018}
}

@inproceedings{greff2022kubric,
  title={{Kubric: A Scalable Dataset Generator}},
  author={Greff, Klaus and Belletti, Francois and Beyer, Lucas and Doersch, Carl and Du, Yilun and Duckworth, Daniel and Fleet, David J and Gnanapragasam, Dan and Golemo, Florian and Herrmann, Charles and others},
  booktitle={Proceedings of the IEEE/CVF conference on computer vision and pattern recognition},
  pages={3749--3761},
  year={2022}
}

@inproceedings{yu2024vonet,
title={{VONet: Unsupervised Video Object Learning With Parallel U-Net Attention and Object-wise Sequential VAE}},
author={Haonan Yu and Wei Xu},
booktitle={The Twelfth International Conference on Learning Representations},
year={2024},
}

@InProceedings{aydemir2023solv,
    author = {Aydemir, G\"orkay and Xie, Weidi and G\"uney, Fatma},
    title = {{Self-Supervised Object-Centric Learning for Videos}},
    booktitle = {Advances in Neural Information Processing Systems},
    year      = {2023}
}

@inproceedings{zhao2023ocmot,
  title={{Object-Centric Multiple Object Tracking}},
  author={Zhao, Zixu and Wang, Jiaze and Horn, Max and Ding, Yizhuo and He, Tong and Bai, Zechen and Zietlow, Dominik and Simon-Gabriel, Carl-Johann and Shuai, Bing and Tu, Zhuowen and others},
  booktitle={Proceedings of the IEEE/CVF international conference on computer vision},
  pages={16601--16611},
  year={2023}
}

@article{wu2022slotformer,
  title={{SlotFormer: Unsupervised Visual Dynamics Simulation with Object-Centric Models}},
  author={Wu, Ziyi and Dvornik, Nikita and Greff, Klaus and Kipf, Thomas and Garg, Animesh},
  journal={International Conference on Learning Representations},
  year={2023}
}

@article{cavanagh2011visual,
  title={{Visual Cognition}},
  author={Cavanagh, Patrick},
  journal={Vision Research},
  volume={51},
  number={13},
  pages={1538--1551},
  year={2011},
  publisher={Elsevier}
}

@article{uijlings2013selectivesearch,
  title={{Selective Search for Object Recognition}},
  author={Uijlings, Jasper RR and Van De Sande, Koen EA and Gevers, Theo and Smeulders, Arnold WM},
  journal={International Journal of Computer Vision},
  volume={104},
  pages={154--171},
  year={2013},
  publisher={Springer}
}

@inproceedings{caron2021dinov1,
  title={{Emerging Properties in Self-Supervised Vision Transformers}},
  author={Caron, Mathilde and Touvron, Hugo and Misra, Ishan and J{\'e}gou, Herv{\'e} and Mairal, Julien and Bojanowski, Piotr and Joulin, Armand},
  booktitle={Proceedings of the IEEE/CVF international conference on computer vision},
  pages={9650--9660},
  year={2021}
}

@article{oquab2024dinov2,
  title={{DINOv2: Learning Robust Visual Features without Supervision}},
  author={Oquab, Maxime and Darcet, Timoth{\'e}e and Moutakanni, Th{\'e}o and Vo, Huy and Szafraniec, Marc and Khalidov, Vasil and Fernandez, Pierre and Haziza, Daniel and Massa, Francisco and El-Nouby, Alaaeldin and others},
  journal={Transactions on Machine Learning Research Journal},
  year={2024}
}

@inproceedings{zhao2025dias,
  title={{Slot Attention with Re-Initialization and Self-Distillation}},
  author={Zhao, Rongzhen and Zhao, Yi and Kannala, Juho and Pajarinen, Joni},
  booktitle={ACM Multimedia},
  year={2025}
}

@inproceedings{seitzer2023dinosaur,
  title={{Bridging the Gap to Real-World Object-Centric Learning}},
  author={Seitzer, Maximilian and Horn, Max and Zadaianchuk, Andrii and Zietlow, Dominik and Xiao, Tianjun and Simon-Gabriel, Carl-Johann and He, Tong and Zhang, Zheng and Sch{\"o}lkopf, Bernhard and Brox, Thomas and others},
  booktitle={The Eleventh International Conference on Learning Representations},
  year={2023}
}
